\documentclass[runningheads]{llncs}
\usepackage{graphicx}
\usepackage{capt-of}
\usepackage{float}
\usepackage[hidelinks]{hyperref}
\begin{document}
\title{Can AI assistance aid in the grading of handwritten answer sheets?}
%
%
\author{Pritam Sil
\and
Parag Chaudhuri
\and
Bhaskaran Raman
}
\authorrunning{Sil et al.}
%
\institute{IIT Bombay, India \\
\email{\{pritamsil,paragc,br\}@cse.iitb.ac.in}}
\maketitle              
\begin{abstract}
With recent advancements in artificial intelligence (AI), there has been growing interest in using state of the art (SOTA) AI solutions to provide assistance in grading handwritten answer sheets. While a few commercial products exist, the question of whether AI-assistance can actually reduce grading effort and time has not yet been carefully considered in published literature. This work introduces an AI-assisted grading pipeline. The pipeline first uses text detection to automatically detect question regions present in a question paper PDF. Next, it uses SOTA text detection methods to highlight important keywords present in the handwritten answer regions of scanned answer sheets to assist in the grading process. We then evaluate a prototype implementation of the AI-assisted grading pipeline deployed on an existing e-learning management platform. The evaluation involves a total of 5 different real-life examinations across 4 different courses at a reputed institute; it consists of a total of 42 questions, 17 graders, and 468 submissions. We log and analyze the grading time for each handwritten answer while using AI assistance and without it.  Our evaluations have shown that, on average, the graders take 31\% less time while grading a single response and 33\% less grading time while grading a single answer sheet using AI assistance.
\end{abstract}
\section{Introduction}
Grading of handwritten answer sheets is a time consuming process. Traditionally, during the grading of handwritten answer sheets the grader has to go through each answer and grade it without any external assistance. On the other hand, recent advances in AI have shown a growing interest in using SOTA methods to provide assistance in this area.  Presently, most commercial solutions today provide an easy-to-use user interface (UI) that the graders can use to upload their set of answer sheets, add rubrics and grade them. While some strictly restrict the question paper format and answer format in order to provide assistance, others relax these restrictions by only providing an easy-to-use UI (more has been discussed in Section~\ref{Related_Work}). Moreover, none of these solutions provide AI-based assistance in grading short and long-answer type questions by highlighting certain keywords that will help them in grading such answers. 
\par
Hence, this work has designed and deployed a prototype implementation of an AI-assisted grading pipeline (details in Section~\ref{AI_Assisted_Grading}) on an existing e-learning platform and verified if keyword highlighting is effective in providing assistance to the grading process. The main contributions of this work are as follows
\begin{itemize}
    \item Highlighting of important keywords in the answer regions of scanned answer sheets to enable faster grading. Question regions are auto-detected on the question paper. 
    \item An extensive field evaluation on over 5 different real-life examinations across 4 different courses at a reputed institute, consisting of a total of 42 questions, 17 graders, and 468 submissions.
      
    \item Our evaluations show that on average, there was a \textbf{31\%} reduction in grading time for a single response and a \textbf{33\%} reduction in grading time for a single answer sheet using AI assistance (as discussed in Section~\ref{Results}).
    
\end{itemize}
\vspace{-0.4cm}
\section{Related Work}
\label{Related_Work}
\vspace{-0.6cm}
\begin{table}[h!tb]
    \centering
    \begin{tabular}{||p{3cm}|p{1.2cm}|p{1.3cm}|p{2cm}|p{2cm}|p{2cm}||}
    \hline
     & \textbf{GS}  & \textbf{SAFE} & \textbf{DG} & \textbf{SG}  & \textbf{Our method}\\
         \hline\hline
       Permitted Answer Types & Any & Any & Textual & A small phrase or word  & Any \\ \hline
Question Paper Format & Any & Any & Any & Restricted  & Any\\ \hline
Auto-grouping of Similar Answers& Y & N & N & N & N \\ \hline
Auto Detection of  Question Regions& N & N & N & N & \textbf{Y} \\ \hline
Auto Extraction of Roll Number& Y & N & N & Y  & Y \\ \hline
Keyword highlighting & N & N & N & N & \textbf{Y} \\ \hline
Field evaluation of methodology & Opinion survey & Opinion based & Not available publicly & Not available publicly  & \textbf{Grading time based} \\
         \hline
    \end{tabular}
    \\
    \caption{Comparison of existing methods with our method}
    \label{related_work}
\end{table}
\vspace{-0.7cm}
The logistics involved in grading handwritten answer sheets can be quite challenging for courses with large enrollments. Gradescope~\cite{b1} (GS) by Singh et al. allows the grader to manually demarcate the answer and roll number regions on a blank question paper, upload a set of scanned answer sheets and subsequently use this to highlight the whole answer region on the answer sheet. Gradescope supports AI based grouping of similar answers.  However, the system does not crop out the answer regions; instead, it just places a bounding box to highlight the answer region for a question as demarcated by the instructor. Additionally, it does not provide any assistance while grading paragraph-type answers.

Smart Authenticated Fast Exams (SAFE) by Chebrolu et al. \cite{b2}  provides a secure solution to conduct examinations in a proctored environment while following a Bring Your Own Device (BYOD) model. SAFE provides a one stop solution for managing day-to-day class activities. The SAFE system allows upload of answer sheet images and provides a manual grading interface for the graders to correct such uploads. However, it does not assist in grading such images, like our pipeline. 
 
Smartail.ai came up with their flagship product called DeepGrade~\cite{b4} (DG). DeepGrade can generate a question paper automatically from an instructor created question set. The students upload their answers to the system in the form of text or handwritten answers. For handwritten answers, the interface will automatically perform optical character recognition(OCR) and automatically grade these answers. DeepGrade is effective in grading short answers where the students have to write a short phrase or a short paragraph. The interface provides some feedback on the grammar and spelling errors present in the answer. However, the system does not automatically detect the question regions, or highlight important keywords present in handwritten answers. 

Swift Grade~\cite{b5} (SG) is a solution for assisting in the grading of answer sheets. Their interface allows the instructor to add questions along with their marks and answers and also register students for a course. The interface generates an answer sheet with the student's names on top, which the instructor can download and distribute to the students during the examination. The interface records which bounding box represents the answer to which question. The answer sheets are scanned and uploaded to the system and are automatically graded. Their interface can automatically grade short equations, phrases and numbers. However, Swift Grade requires the grader to use the generated answer sheet format and does not provide much assistance in grading short and long answer type questions.

Unlike some of the others, ours is a human-in-the-loop solution to answer script grading. We believe while there is merit in providing assistance to the task of grading handwritten answers, doing it completely automatically may not be able to assess the variety of answers possible. We observe in our analysis that error or noise in the output of AI models can adversely the effect performance of such systems, hence it is always better to have a human in the loop, so as to not pass on this adverse effect to the students being assessed.

Table~\ref{related_work} compares various aspects of  GradeScope, DeepGrade, Swift Grade, SAFE, and our AI-assisted grading pipeline. Regarding field evaluation, only  Gradescope and SAFE have made their data publicly available. While Gradescope performed an opinion-based survey and an analysis of grading time at assignment level, SAFE only performed an opinion-based survey. DeepGrade and Swift Grade do not have any publicly available field evaluation data. Our system records the grading time at per question level, and we analyse it as a part of our field evaluation. 

\section{AI Assisted Grading Pipeline}
\label{AI_Assisted_Grading}
\begin{figure}[h!tb]
  \centering
  \includegraphics[scale=0.07]{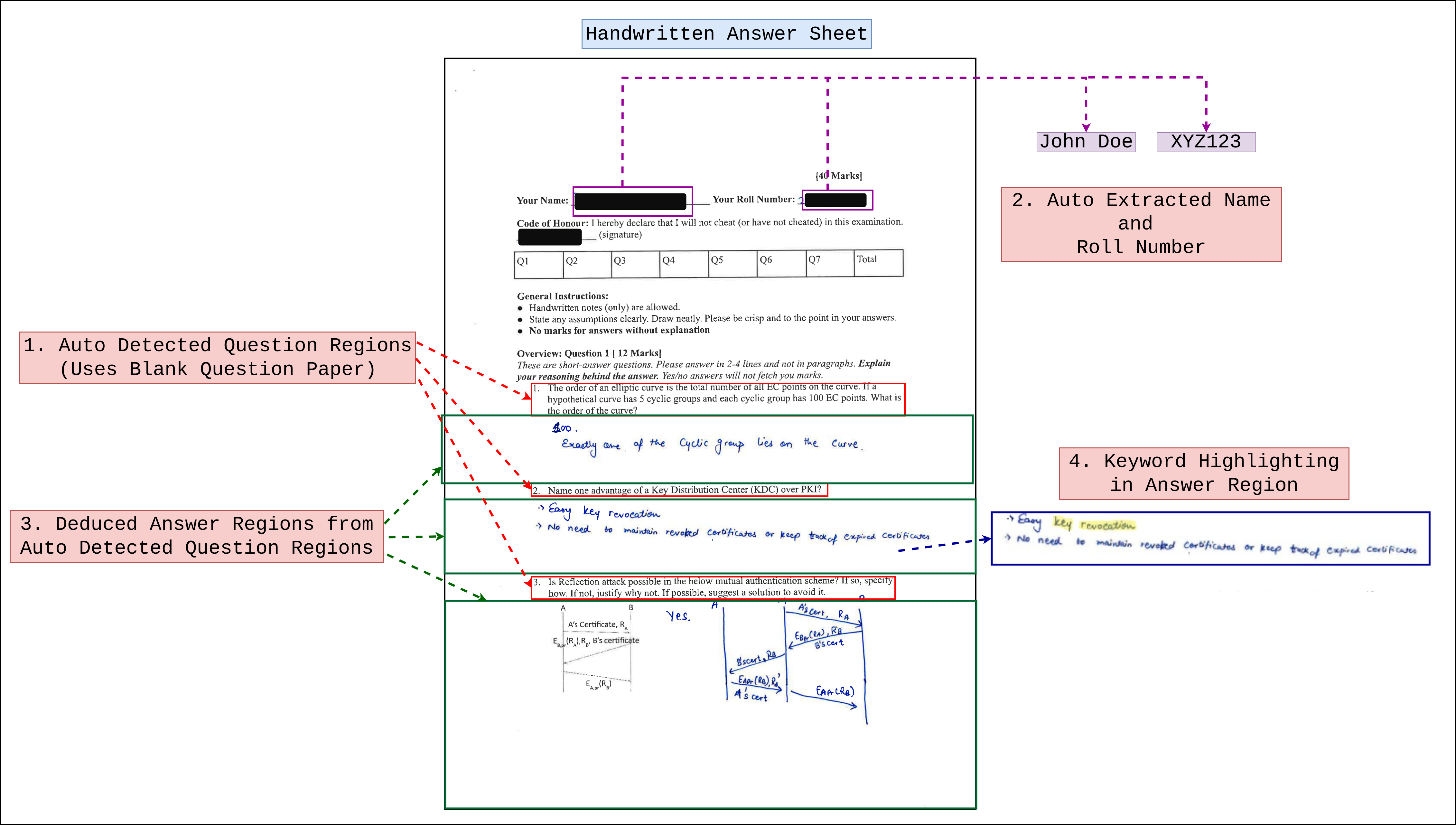}
  \caption{Overview of AI assistance provided by our pipeline}
  \label{overview}
\end{figure}
 The AI-assisted grading pipeline, as shown in Figure~\ref{overview}, assumes that each question will have an answer region provided just below it, and the students will write their answers within this region. Also, the blank question paper must be available as a single PDF file at the start of the process.
The first part of this pipeline automatically detects the question regions 
(\cite{b6}\cite{b11})
present in the PDF file of a blank question paper uploaded by the instructor. Once the instructor is satisfied with the detected question bounding boxes and the corresponding text, they can be saved to the interface. 
 
 Once the examination is over, the instructor will have a set of scanned handwritten answer sheets. Note each such answer sheet is scanned to a separate PDF file. It is difficult to manually create a mapping between each such scanned PDF file to the corresponding student's name and roll number. Our pipeline partially automates this problem using AI. The instructor uploads this set of scanned answer sheets to our interface, and also marks the corresponding bounding box for name and roll number on one PDF. Once done, the system will automatically extract (perform word recognition) the text present in these bounding boxes. For this word recognition step, we use the docTR library as described earlier for question text extraction.
Once each answer sheet has been mapped to a student's name and roll number, the mappings are shown to the instructor via a separate user interface. Incorrect mappings, if any, can be manually corrected by the instructor. Once done, the mappings are saved.

Now, the system automatically deduces the handwritten answer regions from each answer sheet. The deduced answer region for a particular question is the region between the bottom part of the bounding box of that question and the top part of the bounding box of the next question. The system automatically crops out the handwritten answer region for a question for a particular student's answer sheet and displays it on the grading interface. 
This allows the grader to focus on the relevant region where the answer is present. The grader has the option to look at the entire page image, if needed.

It is often seen that graders look for specific keywords while grading a particular answer. However, as seen in see Section ~\ref{Related_Work}, this has not been investigated by any earlier or currently available system. Our system asks the instructor to input relevant keywords for each answer once in the process and subsequently highlights\cite{b10} them on the handwritten answer regions for a particular question. 

\section{Field evaluation of grading time reduction}
\label{Results}
This section describes the field evaluation of the AI-assisted grading pipeline described above.

\subsection{Evaluation Method}
A prototype implementation of the pipeline is used to grade real-life examinations at a reputed institute\footnote{Name withheld for anonymity}. The handwritten answer sheets of the students are scanned and uploaded to the platform, and the answer regions were highlighted using the given keywords. 
The graders were then asked to grade their respective questions using the manual grading interface, and the time needed to grade each response was automatically recorded by the system. No timer was shown to the grader, but the time was logged by the system in a manner transparent to the grader. This was done to ensure a fair evaluation and not put any pressure on the grader. 

\begin{minipage}{\textwidth}
  \begin{minipage}[b]{0.49\textwidth}
    \centering
    \begin{tabular}{||p{1.5cm}|p{1cm}|p{1.2cm}|p{1.3cm}||}
    \hline
      \textbf{Course}   & \textbf{\# Sub} & \textbf{\# Qs} &  \textbf{\# Graders}\\
         \hline\hline
       Course A & 124  &  21 & 3\\
         \hline
       Course B &  49  & 10  & 3\\
         \hline
       Course C & 50  &  1 & 1\\
         \hline
        Course D  & 198  &  5 & 9 \\
         \hline
        Course E  & 47  &  5 & 1 \\
         \hline
    \end{tabular}
    \captionof{table}{Details of each examination}
    \label{exam}
  \end{minipage}
  \hfill
  \begin{minipage}[b]{0.49\textwidth}
      \centering
      \includegraphics[scale=0.16]{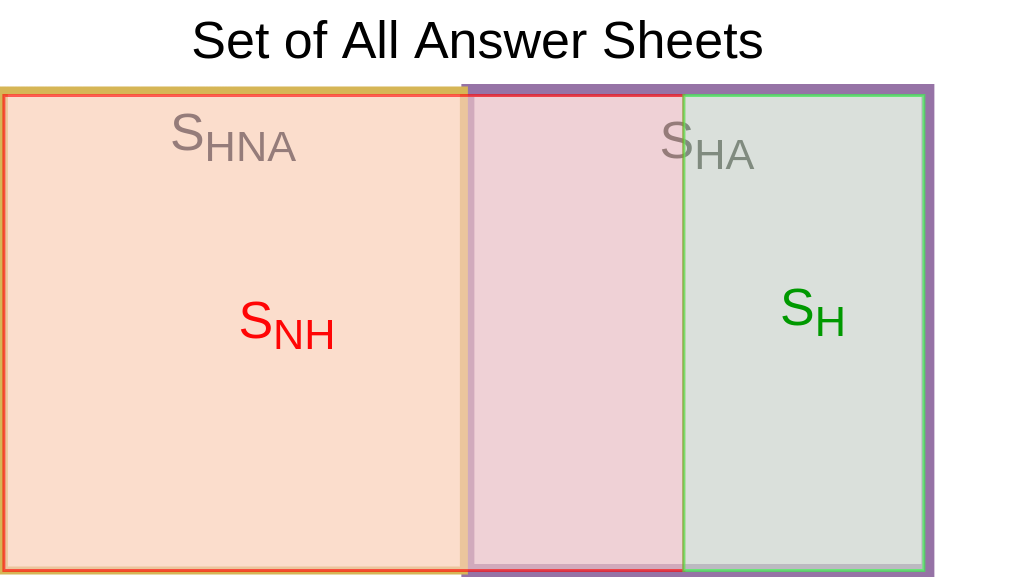}
      \captionof{figure}{Grouping of Answer Sheets\\ for field evaluation}
      \label{khighlightset}
    \end{minipage}
\end{minipage}
  
\ \\
The pipeline was evaluated on five examinations for five separate courses. Table~\ref{exam} displays the number of submissions and the number of questions in which keyword highlighting has been used for each examination. It can be seen that the evaluation dataset is varied and has a significant number of answer sheets. Grading all of them manually would have taken a non-trivial amount of time. Each of these examinations contains a significant number of questions, except the Course C examination, which had only 1 question that could use the keyword highlight feature. Also, all these questions are of different types, including those with numerical answers, short text answers and long paragraph answers. 
The set of scanned answer sheets was split into two equal parts (see Figure~\ref{khighlightset}). In the first half, no attempt was made to highlight keywords present in it ($S_{HNA}$). In the second half, an attempt was made to highlight keywords in it ($S_{HA}$). Since no handwritten text detector and recognizer is $100\%$ accurate, certain images in $S_{HA}$ did not have any keyword highlighted within them. These are contained in the set $S_{NH}$(the red rectangle). $S_{H}$ (the green rectangle) denotes the set of answer sheets where at least one keyword was highlighted. Thus, due to the inaccuracy of the keyword detection model, $S_{HA}$ contains elements from both $S_{H}$ and $S_{NH}$. 

The main goal of the evaluation was to determine whether graders take less time to grade the student's answer or not when using keyword-highlighting feature\footnote{The keyword highlight feature is used interchangeably with AI-based assistance in this section.}. The time taken to grade each response is recorded and analyzed. It should be noted that \textit{no personally identifiable student information was used} for the study.

\subsection{Analysis at per response level (Course B)}
During our evaluation process, the time taken to grade each response was recorded. Both the $S_{HNA}$ and $S_{HighlightAttempted}$ sets contained few grading time instances, which are quite high compared to the mean time recorded. We conjecture that these maybe cases where the grader took a break between grading one question and the next. These are considered to be outliers and are eliminated by removing the top $5$\% data points in a set.  
    \begin{figure}[h!tb]
      \centering
      \includegraphics[scale=0.135]{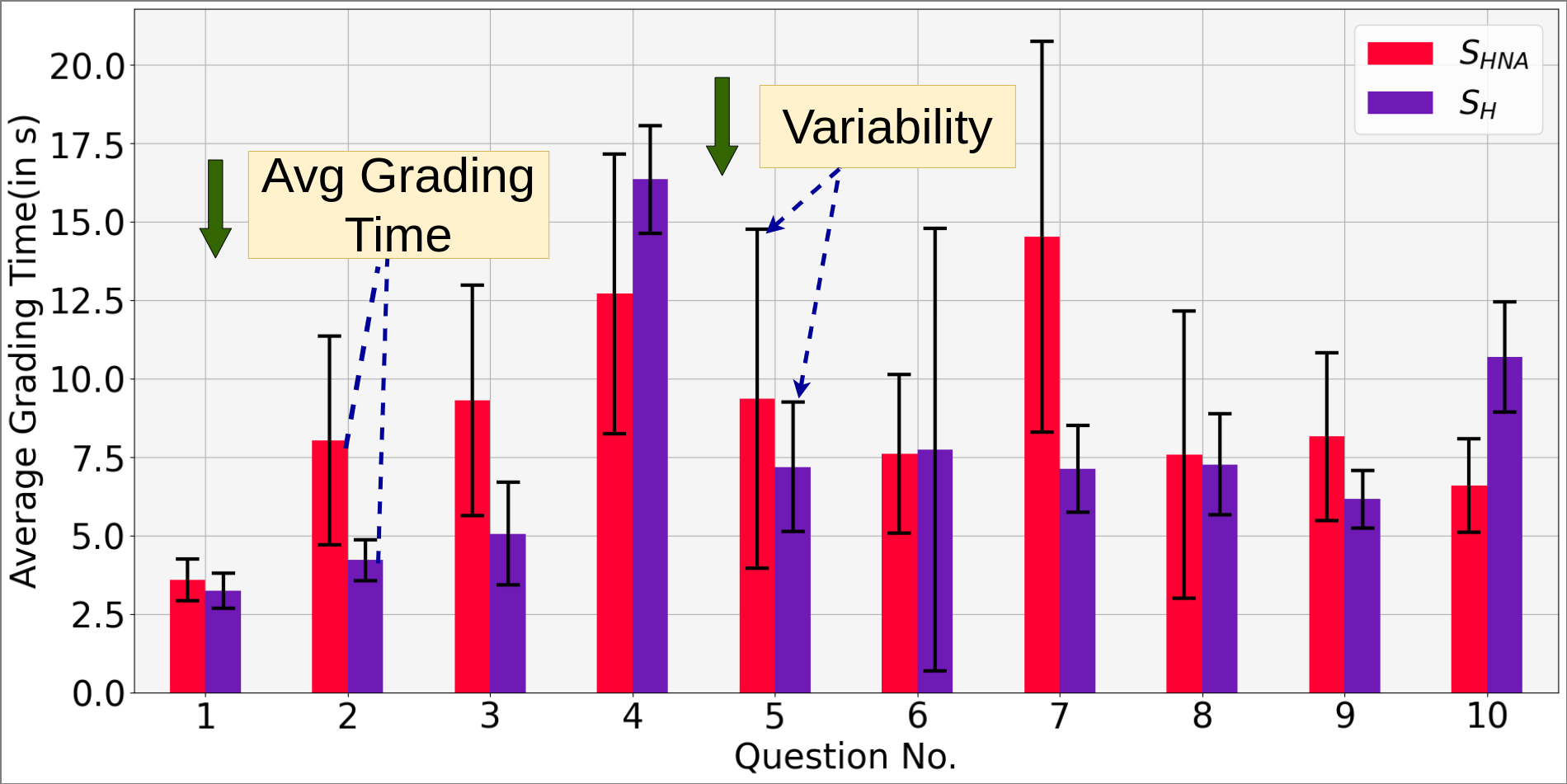}
      \caption{Average grading time ({S\textsubscript{HNA}} vs {S\textsubscript{H}}) for Course B}
      \label{Course B_noerror}
    \end{figure}   
Figure~\ref{Course B_noerror} shows the average grading time, for both $S_{HNA}$ and $S_{H}$ for each question present in Course B.  It can be seen that the graders have taken less time to grade all the questions using AI-based keyword highlighting, except questions 4, 6 and 10. Question 4 asks the students to write pseudocode as a solution. It is thus seen that highlighting keywords in pseudocode is not effective in reducing the grading time. Also, for questions 4 and 6 involved the grader going through the entire explanation provided by the student. This suggests that for questions where the logic present within the answer is more important than looking at certain keywords, the keyword highlighting feature cannot assist the grader. 

We also observe in Figure~\ref{Course B_noerror} that the variability in grading time is significantly reduced while using AI assistance. It is often seen that student's answers are needlessly verbose. We conjecture that because of the keyword highlight feature, the grader can focus on the specific regions within the answer where the relevant parts of the explanation are present. Thus, for each answer the grader has to go through a specific set of sentences rather than the whole answer. Hence, this reduces the variability in grading time while using the keyword feature as each time the grader has to go through approximately same number of lines.

While we have presented results here for one of the five examinations in detail, we note that similar results were found for the other examinations. To summarize this section, we compute the average of individual reductions in grading time at the response level for each question from this data (as shown in Figure~\ref{Course B_noerror} and similar data from other examinations). We compute average reduction in grading time by computing the average of reductions of grading times of each question w.r.t $S_{HNA}$. The average reduction in grading time per response as computed for Q is $31\%$. (More Details :  \href{https://sites.google.com/view/pritam-sil/aied2024}{\underline{https://sites.google.com/view/pritam-sil/aied2024}})

\subsection{Analysis at per answer sheet level}
The time to grade an answer sheet is considered to be the total time taken to grade all the questions in that answer sheet. The average grading per answer sheet for each examination is analysed here.
\begin{figure}[h!tb]
  \centering
  \includegraphics[scale=0.1]{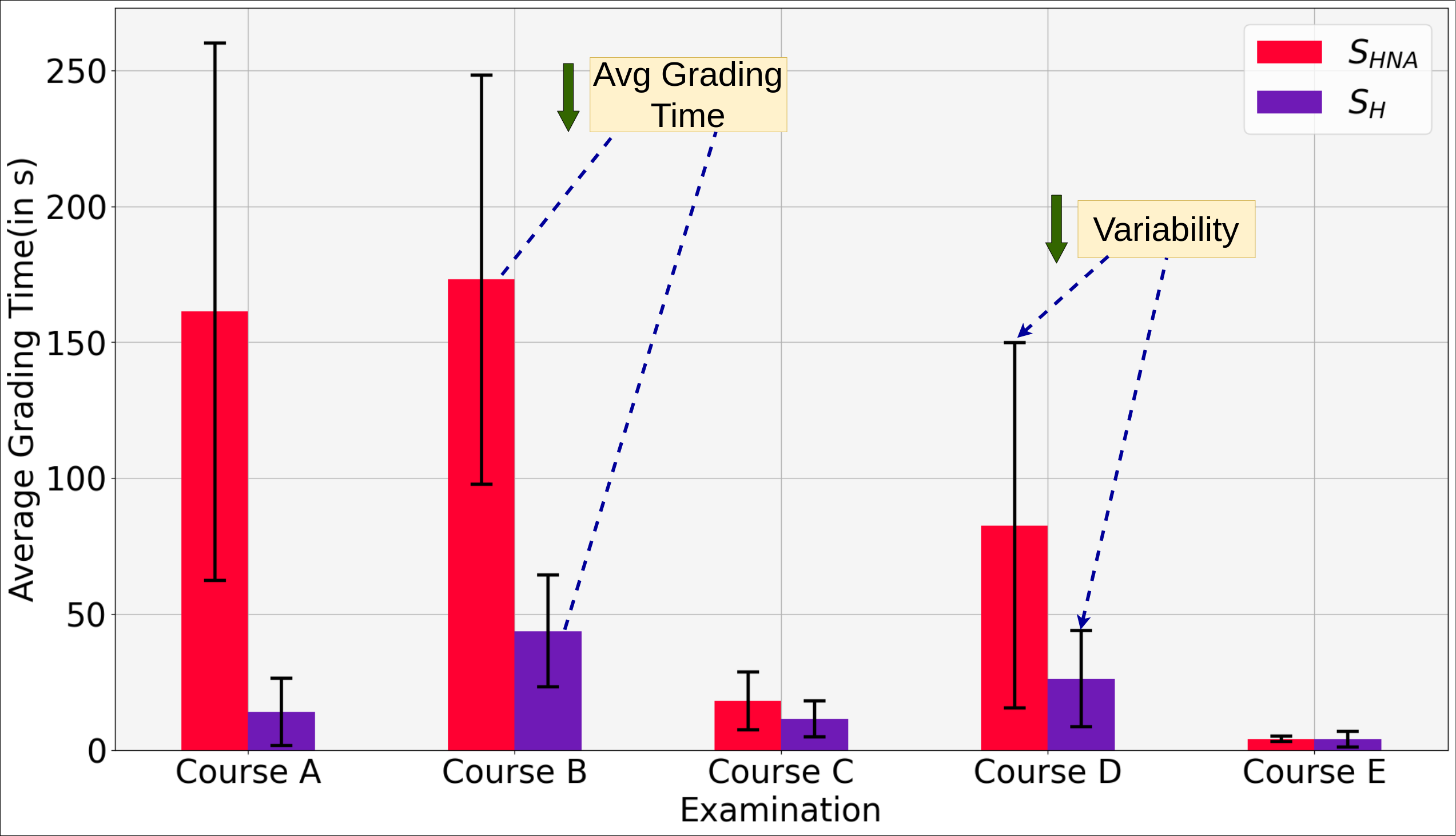}
  \caption{Average grading time - {S\textsubscript{HNA}} vs {S\textsubscript{H}}(Answer Sheet)}
  \label{overall_noerror}
\end{figure}
Figure~\ref{overall_noerror} shows the average grading time per answer sheet across all examinations, for both $S_{HNA}$ and  $S_{H}$. It clearly shows the fact that graders took less time while grading the whole answer sheet while using keyword highlighting than when they did not have that assistance. 

Overall variability in average grading time can again be observed to be low in Figure~\ref{overall_noerror} when graders used keyword highlighting. Computing the average of individual reduction in average grading time at the answer sheet level for each examination from this data (as shown in Figure~\ref{overall_noerror}) indicates an overall reduction of $33\%$ in grading time per answer sheet. Similar analysis at question type level revealed a reduction of 23\% for long answer type questions , 34\% for numerical type and 34\% for short answer type questions. 
\vspace{-0.3cm}
\section{Conclusion}
\vspace{-0.2cm}
We introduce an AI-assisted pipeline for grading of handwritten answer scripts. The system aids the grader by automatically detecting the question regions and text in the blank question paper. Subsequently, it extracts the name and roll number of students from scanned answer sheets and maps the scanned answer sheets to corresponding students. At grading time, it automatically crops out the handwritten answer regions and displays it on our manual grading interface and highlights important keywords in the answer. Extensive field evaluation of this pipeline reveals a $31\%$ reduction in average grading time for each handwritten response and a $33\%$ reduction in average grading time for entire answer sheets, when using the keyword highlight feature. Thus, AI-based assistance can definitely aid in the grading of handwritten answer sheets.

The current work only highlights keywords if there is an exact match. Future work involves but is not limited to using techniques from Natural Language Processing (NLP) to detect different forms or even synonyms of the keywords and eventually automatically grade such handwritten answers.
\vspace{-0.3cm}
\section*{Acknowledgements}
\vspace{-0.1cm}
We would like to thank the TIH Foundation for IoT and IoE at IIT Bombay for funding this project (RD/0121-TIH0000-001).

\bibliographystyle{splncs04}
\bibliography{mybib}

\begin{thebibliography}{1}
\providecommand{\url}[1]{\texttt{#1}}
\providecommand{\urlprefix}{URL }
\providecommand{\doi}[1]{https://doi.org/#1}

\bibitem{b4}
{DeepGrade}. \url{https://smartail.ai/deepgrade/}, last accessed: 2024-01-25

\bibitem{b11}
{docTR}. \url{https://mindee.github.io/doctr/}, last accessed: 2024-01-25

\bibitem{b6}
{pdfminer}. \url{https://pdfminersix.readthedocs.io/}, last accessed:
  2024-01-25

\bibitem{b5}
{Swift Grade}. \url{https://goswiftgrade.com/}, last accessed: 2024-01-25

\bibitem{b2}
Chebrolu, K., Raman, B., Dommeti, V.C., Boddu, A.V., Zacharia, K., Babu, A.,
  Chandan, P.: {SAFE: Smart Authenticated Fast Exams for Student Evaluation in
  Classrooms}. In: Proceedings of the ACM SIGCSE Technical Symposium on
  Computer Science Education. p. 117–122. ACM, New York, NY, USA (2017)

\bibitem{b10}
Kuang, Z., Sun, H., Li, Z., Yue, X., Lin, T.H., Chen, J., Wei, H., Zhu, Y.,
  Gao, T., Zhang, W., Chen, K., Zhang, W., Lin, D.: {MMOCR:} {A} comprehensive
  toolbox for text detection, recognition and understanding. CoRR
  \textbf{abs/2108.06543} (2021)

\bibitem{b1}
Singh, A., Karayev, S., Gutowski, K., Abbeel, P.: Gradescope: A fast, flexible,
  and fair system for scalable assessment of handwritten work. In: Proceedings
  of the Fourth (2017) ACM Conference on Learning @ Scale. pp. 81--88 (04 2017)

\end{thebibliography}
\end{document}